\documentclass[letterpaper, 10 pt, conference]{ieeeconf}  

\IEEEoverridecommandlockouts                              

\usepackage{graphics} 
\usepackage{amsmath} 
\usepackage{amssymb}  
\usepackage{todonotes}
\usepackage[nolist]{acronym}
\usepackage{algorithm}
\usepackage{algorithmic}
\usepackage{svg}
\usepackage{subfigure}
\usepackage{siunitx}
\presetkeys{todonotes}{}{}
\usepackage{amsmath}
\usepackage{caption}
\usepackage{cite}
\usepackage{bm}
\usepackage{booktabs} 
\usepackage{hyperref}

\usepackage[normalem]{ulem}
\definecolor{OliveGreen}{RGB}{0,200,25}
\newcommand{\red}[1]{\textcolor{red}{#1}}
\newcommand{\darkgreen}[1]{\textcolor{OliveGreen}{#1}}

\newcommand{\replaced}[2]{\red{\ifmmode\text{\sout{\ensuremath{#1}}}\else\sout{#1}\fi} \darkgreen{#2}}
\newcommand{\deleted}[1]{\red{\ifmmode\text{\sout{\ensuremath{#1}}}\else\sout{#1}\fi}}

\def\eqref#1{equation~\ref{#1}}

\def\1{\bm{1}}

\DeclareMathAlphabet{\mathsfit}{\encodingdefault}{\sfdefault}{m}{sl}
\SetMathAlphabet{\mathsfit}{bold}{\encodingdefault}{\sfdefault}{bx}{n}

\title{\LARGE \bf
ETA-IK: Execution-Time-Aware Inverse Kinematics \\ for Dual-Arm Systems
}

\author{*Yucheng Tang$^{1,2,3}$, *Xi Huang$^{2}$, Yongzhou Zhang$^{1,2}$, Tao Chen$^{2,3}$, Ilshat Mamaev$^{1,3}$, Bj\"orn Hein$^{1,2}$
\thanks{*This paper is the scientific result of a research project "ROBDEKON II", funded by the \ac{BMBF}, Grant No. 13N16543.}
\thanks{$^{1}$Karlsruhe University of Applied Sciences, Karlsruhe, Germany }%
\thanks{$^{2}$Karlsruhe Institute of Technology, Karlsruhe, Germany}
\thanks{$^{3}$Proximity Robotics \& Automation GmbH}%
\thanks{*These authors contributed equally to this work. }%
}

\begin{acronym}
\acro{PTP}{point-to-point}
\acro{OTG}{Online Trajectory Generation}
\acro{TG}{Trajectory Generation}
\acro{PSO}{Particle Swarm Optimization}
\acro{DoF}{Degrees of Freedom}
\acro{SDM}{Software-Defined Manufacturing}
\acro{TCP}{Tool Center Point}
\acro{MPC}{Model Predictive Control}
\acro{IK}{Inverse Kinematics}
\acro{SDF}{Signed Distance Field}
\acro{NBV}{Next-Best-View}
\acro{MRTA}{Multi-Robot Task Allocation}
\acro{SDM}{Software-Defined Manufacturing}
\acro{MRS}{Multi-Robot System}
\acro{multi-EE}{Multi-End-Effectors}
\acro{EE}{End-Effector}
\acro{BMBF}{German Federal Ministry of Education and Research}
\end{acronym}

\begin{document}

\maketitle
\thispagestyle{empty}
\pagestyle{empty}

\begin{abstract}
This paper presents ETA-IK, a novel Execution-Time-Aware Inverse Kinematics method tailored for dual-arm robotic systems. The primary goal is to optimize motion execution time by leveraging the redundancy of the entire system, specifically in tasks where only the relative pose of the robots is constrained, such as dual-arm scanning of unknown objects. Unlike traditional IK methods using surrogate metrics, our approach directly optimizes execution time while implicitly considering collisions. A neural network based execution time approximator is employed to predict time-efficient joint configurations while accounting for potential collisions.
Through experimental evaluation on a system composed of a UR5 and a KUKA iiwa robot, we demonstrate significant reductions in execution time. The proposed method outperforms conventional approaches, showing improved motion efficiency without sacrificing positioning accuracy.
\end{abstract}

\section{INTRODUCTION}
\label{sec:introduction}

\begin{figure}[tbp]
    \centering
    \includegraphics[width=0.48\textwidth]{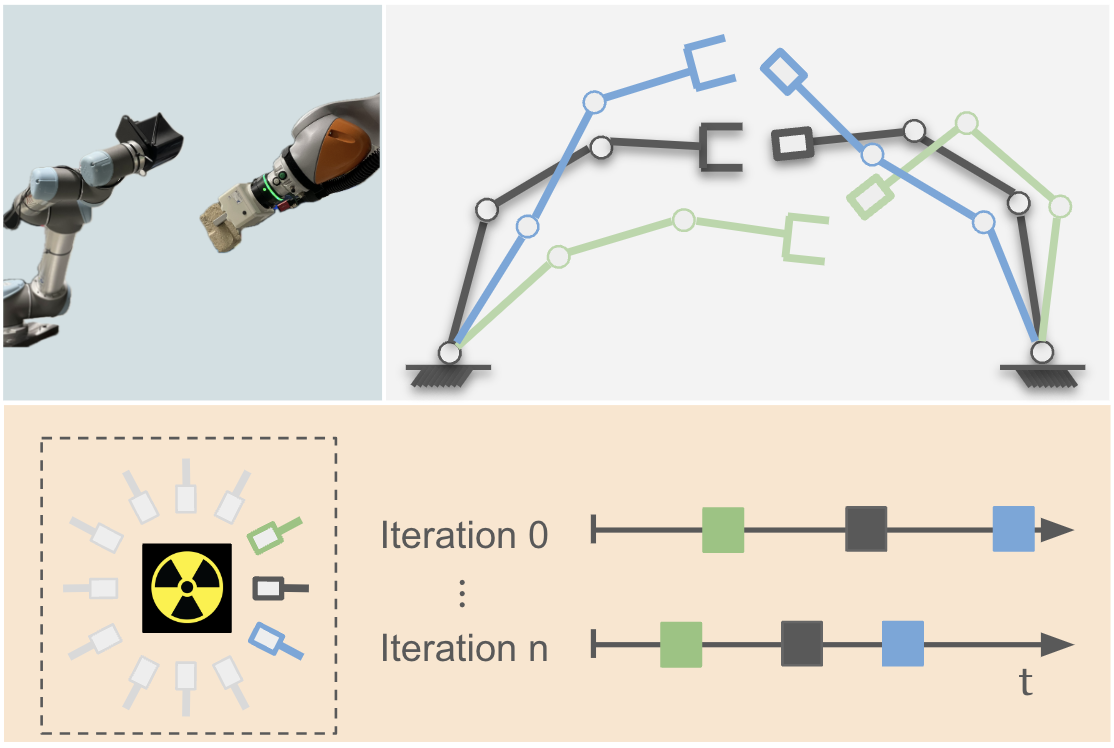}
    \caption{\textbf{Application and Motivation}: Improving execution time for modeling radioactive unknown objects with two robot arms (top left). Given a set of relative poses representing the perception poses (bottom left), ETA-IK aims to use the redundant \ac{DoF} to solve each single IK problem and find the optimal and collision-free joint configurations (top right), to accelerate the entire scanning process (bottom right). }
    \label{fig:pipeline1}
\end{figure}

While robots are becoming increasingly capable of performing everyday tasks in Human-Robot Collaboration settings, there are still many scenarios where robots must operate autonomously, without humans present. Notable examples include the dismantling of nuclear facilities, disaster relief operations, and tasks conducted in hazardous environments \cite{petereit_robdekon_2019}. In these environments, factors such as human-like motion and predictable behavior, which are crucial for human-robot collaboration, become secondary. Instead, execution efficiency and collision avoidance become the primary concerns. As shown in Fig. \ref{fig:pipeline1}, this work originates from a real-world scenario involving the scanning and modeling of unknown objects during the dismantling and decontamination of a nuclear facility. Robots must efficiently coordinate to achieve rapid 3D modeling and proceed further manipulation such as radioactive scanning and wiping over the object based on this 3D model. 
Given a sequence of discrete object-centric perception poses (e.g. generated with \ac{NBV} methods \cite{pan_scvp_2022}), this work focuses on finding the joint configurations of a dual-arm system for each pose with an objective to minimize the execution time of the object modeling process. Compared to continuous scanning, discrete perception poses allows for more time-efficient execution, as the robots can move between viewpoints with full acceleration and velocity capabilities. Additionally, the motion between two robotic arms enables faster execution than a single arm, as velocity and acceleration limits can be superimposed across both manipulators. Furthermore, dual-arm systems provide additional \ac{DoF} compared to single-arm setups, enabling more flexible solutions that can reduce motion time, avoid self-collisions, and better adapt to task constraints. 

In dual-arm manipulation, a particular class of problem arises where only \textit{the relative motion} of the \ac{EE} is specified as input, imposing a path constraint on the motion planning problem \cite{doi:10.1177/0278364920988087}.
When only the relative \ac{TCP} pose between the two arms is important, the redundant \ac{DoF} allow for multiple kinematic solutions to achieve the same task. Consider a dual-arm manipulation scenario where one arm holds an object while the other utilizes a LiDAR sensor to collect point cloud data (as shown in Fig. \ref{fig:pipeline1}). In such cases, the selection of the joint configurations for the given desired relative \ac{TCP} poses in Cartesian space determines the execution performance of the task. The motion efficiency is determined not only by minimizing joint displacement but also by avoiding potential collisions and accounting for varying joint dynamic constraints. Poor \ac{IK} selection can result in longer execution times.
Therefore, this paper aims to address a \textit{\ac{PTP} relative target pose} problem, which fundamentally differs from the \textit{continuous relative motion} problem \cite{doi:10.1177/0278364920988087}.

To characterize motion efficiency, we define motion execution time—the time required to move from the current robot configuration to the target pose—as our primary metric. The objective of this work is to compute \textit{\ac{PTP} target joint configurations} that satisfy task constraints while minimizing execution time. This metric can be integrated into a multi-objective inverse kinematics framework.


The main contributions of this work are: 
\begin{itemize}
    \item \textit{A neural network-based execution time approximator} is developed that accounts for potential self-collisions, providing accurate predictions regarding execution time of a collision-free point-to-point motion generated by standard \ac{TG}.
    \item \textit{A multi-objective \ac{IK} method} for dual-arm systems is presented that formulates the problem using relative pose between \ac{TCP}s of the robots, optimizing motion efficiency by fully leveraging the system's redundancy.
    \item The motion efficiency is achieved by directly \textit{integrating the execution time approximator into the \ac{IK} optimization framework}, demonstration significant improvements over  traditional approaches that rely on surrogate metrics like joint configuration distance.
    
    
\end{itemize}


\section{Related Work}

\subsubsection*{Explore Redundancy Resolution}
Numerous studies have leveraged redundancy resolution in seven \ac{DoF} robots to enhance motion optimization \cite{flacco_motion_2012, vahrenkamp_representing_2015, tang_towards_2023}. When the objective is to achieve coordinated relative poses between multiple robotic arms, the nullspace of the relative Jacobian becomes larger, enabling further optimization across multiple sub-tasks or asymmetric tasks \cite{su_manipulability_2019} \cite{tarbouriech_dual-arm_2018}.

Some research has extended the concept of the nullspace by incorporating advanced optimization strategies. For instance, cost functions based on the manipulability of key waypoints have been proposed to define secondary tasks \cite{faroni_global_2016}, thereby optimizing the average manipulability index throughout the task execution. Additionally, by deriving the nullspace into acceleration domain, nullspace-based impedance controller is proposed to handle force interactive tasks \cite{lee_relative_2014} and minimization of motion time under kinematic/dynamic bounds can also be achieved \cite{al_khudir_faster_2018}.

To the best of the author's knowledge, there is currently no research that directly incorporates execution time explicitly into the \ac{IK} optimization problem. Some studies \cite{trutman_globally_2022} \cite{votroubek_globally_2024} that aim to find a global optimum introduce a distance cost function specific to the joint configuration. 
However, achieving a global optimum requires converting the optimization problem into a convex form, which restricts the extensibility of objective function and leads to an exponential increase in computational time as the number of redundant degrees of freedom increases. 

In addition, other related methods include inertial load optimization based on Cartesian spatial direction and the concepts of pseudo-inverse velocity and pseudo-inverse acceleration\cite{al_khudir_faster_2018}. 
Some non-convex multi-objective optimization methods \cite{rakita_relaxedik_2018, rakita_shared_2019} also utilize joint configuration distance as an optimization metric. Furthermore, evolutionary algorithms \cite{wu_t-ik_2021, 8449979} have been combined with optimization to improve convergence speed. Although they mentioned \ac{multi-EE} applications in their work, but these methods typically solve separate \ac{IK} porblems for each \ac{EE}.
And these methods are more suitable for local optimization and path following, which differs from the problem addressed in this work. 

\subsubsection*{Time-Optimal Trajectory Generation}
Traditional methods to optimize the execution time are usually incorporated into the trajectory optimization step, after IK solving and path planning. Many approaches \cite{pham_general_2014, pham_new_2017, 9905530} are typically applied as post-processing steps rather than being integrated into the single-pose \ac{IK} solution. They generally assume that a feasible joint target pose or a series of waypoints is already available, which makes them more appropriate for waypoint-based motion planning, as opposed to addressing single-pose \ac{IK} challenges. 

\subsubsection*{Learning-based Approaches}
Regarding the learning based IK solver, very earlier, the authors in \cite{oyama2001inverse} proposed to use a prediction network to reduce the IK solving time. 
Numerous studies \cite{vu2023machine, bocsi2011learning} have explored various neural network architectures for solving the \ac{IK} problem.

Directly learning an IK solution for redundant robots using an end-to-end neural network approach presents significant challenges. IKNET \cite{bensadoun_neural_2022} demonstrated improvements compared to MLP by decreasing the position error into the centimeter level. Unsupervised learning-based IK is proposed for highly redundant system\cite{stephan_learning_2023}, achieving the similar accuracy. Another important finding from this work is that the neural network solver is capable of implicitly generating collision-free solutions in static scenes when trained on datasets containing obstacle avoidance scenarios. On the other hand, for the \ac{IK} solver based on optimization, the network can be used to estimate some metrics, such as the nullspace parameter \cite{towell_learning_2010}.
This work integrates learning-based methods with optimization has the potential to preserve the accuracy of standard approaches while simultaneously enhancing performance metrics that are challenging to model explicitly.

\section{Problem Statement}


The goal of this work is to find a joint configuration $\bm{q} = [\bm{q_A}, \bm{q_B}] $ of a dual-arm system that results in the desired target relative \ac{TCP} pose $\bm{x_R}$ while minimizing the motion execution time $T$ of a trajectory $\bm q_{0:T} = [\bm q_0, \dots, \bm q_T]$ connecting $\bm q$ and a given initial configuration $\bm{q}_s$, where $\bm{q}_0= \bm{q}_s, \bm{q}_T=\bm{q}$. Additional constraints includes zero initial/final joint velocity $\dot{\bm{q}}_0=\dot{\bm{q}}_T=0$, and robot dynamic and kinematic limits as follow:


\begin{equation}
    \begin{aligned}
        &\bm{q}_{\text{min}} \leq \bm{q}(t) \leq \bm{q}_{\text{max}},  \\
        &\dot{\bm{q}}_{\text{min}} \leq \dot{\bm{q}}(t) \leq \dot{\bm{q}}_{\text{max}}, \\
        &\ddot{\bm{q}}_{\text{min}} \leq \ddot{\bm{q}}(t) \leq \ddot{\bm{q}}_{\text{max}}     \hspace{0.5cm}    \forall t \in [0, T].
    \end{aligned} 
\end{equation}

\section{Approach}

Given a sequence of object-centric SE(3) perception poses, we leverage redundancy of the dual arms to optimize the joint configurations for fast execution. First, we use the relative pose between both arms to formulate an inverse kinematics problem. 
Secondly, we convert this formulation into a parallel multi-objectives problem.
Finally, we train a differentiable time estimator to capture the execution time, which can directly incorporated in the objectives for optimization. 
During the optimization, the joint configurations converge to the target desired pose and sequentially improve the execution time.

\begin{figure}[tbp]
    \centering
    \includegraphics[width=0.49\textwidth]{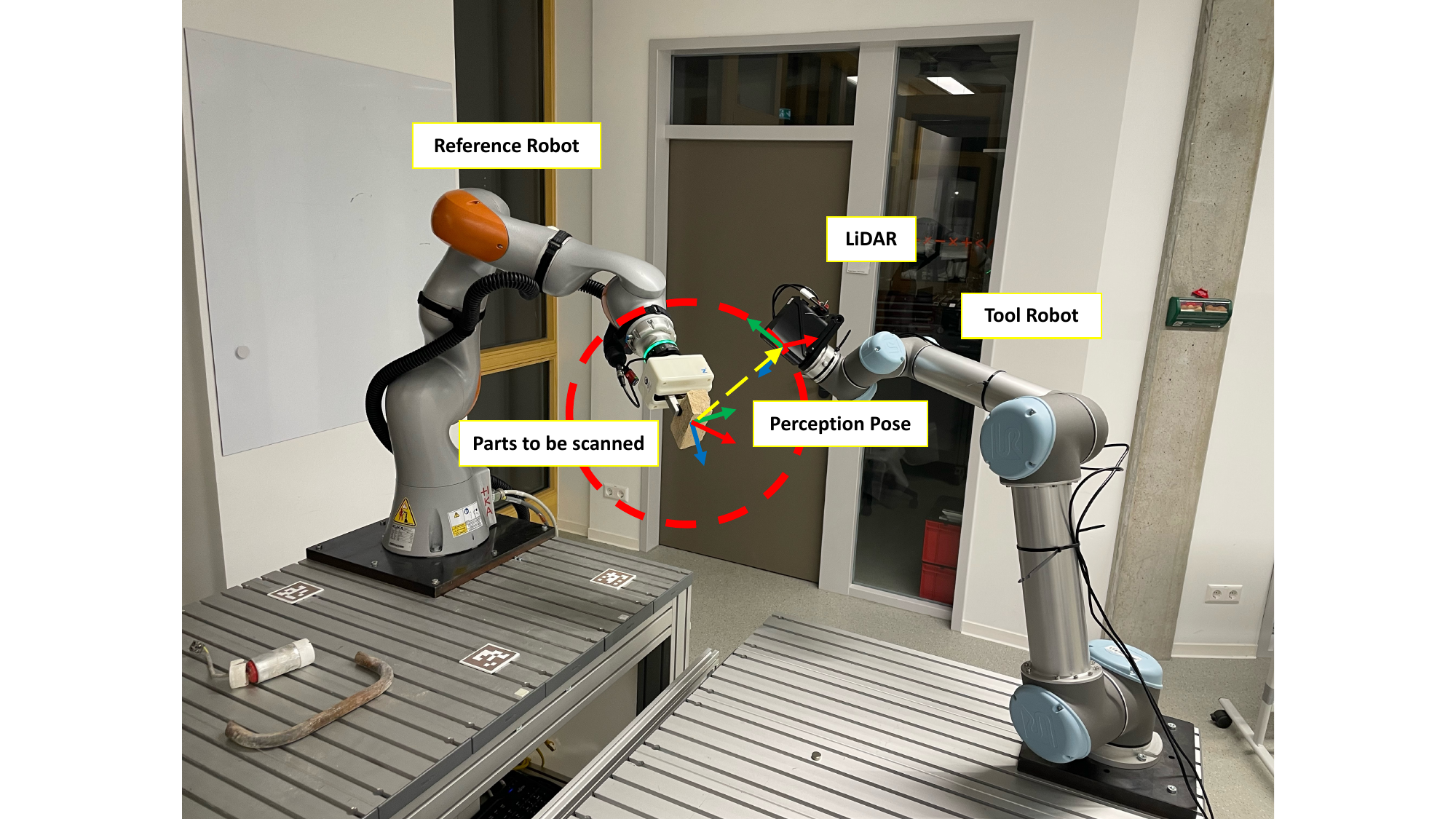}
    \caption{\textbf{Scanning and modeling with two different manipulators}: The KUKA iiwa robot picks up the target object (stone) and the UR5 robot carries a 3D lidar scanner. To get a complete and accurate model, both robots have to change the scanning pose through many relative poses that are projected as the best scanning perspective. This process needs to be accelerated by using our proposed approach.}
    \label{fig:real_robot_experiment_setup}
\end{figure}

\subsection{Relative TCP Pose Approach}\label{sec:relative TCP}

In a dual-arm robot system, as shown in Fig. \ref{fig:real_robot_experiment_setup}, one robot, called the tool robot, holds the tool (camera or scanner), while the other arm, called the reference robot, holds the object to be modeled. The concept of a relative Jacobian is built on vectors that represent the position and orientation of the \ac{TCP} of the tool robot relative to the \ac{TCP} of the reference robot. It imposes a constraint regarding the relative motion between the two \ac{TCP}s without any restrictions regarding their absolute positions. Using this Jacobian, the focus in a two-arm maneuver is the relative motion between the \ac{TCP}.
The relative position vector $\bm{x}_{R}$ is defined as:
\begin{equation}
     \bm{x}_R = \bm{x}_B - \bm{x}_A
\end{equation}
where $\bm{x}_A, \bm{x}_B \in \mathbb{R}^{n_R}$ are the \ac{TCP} pose of the reference robot and the tool robot, respectively and $n_R$ refers to the dimension of robot's workspace. By differentiating this equation with respect to time, we obtain:
\begin{equation}
    \dot{\bm{x}}_R = \dot{\bm{x}}_B - \dot{\bm{x}}_A = J_R \dot{\bm{q}}
\end{equation}
where $\dot{\bm{x}}_R$ represents the relative velocity between two robots' \ac{TCP}s and $\dot{\bm{x}}_B, \dot{\bm{x}}_A$ denote the absolute velocity of the respective robot's \ac{TCP}. 
The relative Jacobian $J_R \in \mathbb{R}^{n_R \times n_T}$ can be expressed in terms of the individual Jacobians of the robots \cite{lee_relative_2014}:
\begin{equation}
    J_R =
    \begin{bmatrix}
      -R_{RA} J_A \\
      R_{RT}^T R_{RB} J_B
    \end{bmatrix}
\end{equation}
where $J_A \in \mathbb{R}^{n_R \times n_A}$ and $J_B \in \mathbb{R}^{n_R \times n_B}$ are the Jacobians of robots A and B, respectively. The term $n_{[\cdot]}$ denotes the dimension of the robot workspace or joint space and $n_T = n_A + n_B$ is the total number of \ac{DoF}s for the system. The block diagonal rotation matrices $R_{RA}, R_{RT}, R_{RB} \in \mathbb{R}^{n_R \times n_R}$ are determined with respect to their associated reference frames. The relative Jacobian $J_R$ provide gradients of the pose error in the multi-objective optimization.

\subsection{Multi-Objective Relative Pose \ac{IK}}
\label{subsec: multi-obj-ik}
The primary objective for relative \ac{IK} problem is to accurately determine the relative pose between the reference robot and the tool robot. To achieve this, the first components of the objective function are defined to address both the relative position and orientation errors.

For the relative position, the objective function can be formulated as:
\begin{equation}
    \chi_p(\bm{q}) = \| \bm{p}_{R,target} - (FK_B(\bm{q}_B) - FK_A(\bm{q}_A) \|_2^2
    \label{eq:position}
\end{equation}
where $FK_A(\bm{q}_A)$ and $FK_B(\bm{q}_B)$ are the forward kinematics functions of robots A and B given their joint configurations, and $\bm{p}_{R,target}$ is the relative target position of the robot B \ac{TCP} with respect to the robot A \ac{TCP}.

For the relative orientation error, the objective function is given by:
\begin{equation}
    \chi_o(\bm{q}) = \text{d}( q_{R,target}, \text{d}(q_B[FK(\bm{q}_B)], q_A[FK(\bm{q}_A)]) ),
    \label{eq:orientation}
\end{equation}
where $d(\cdot)$ is the displacement between two quatenions, $q_{R,target}$ is the relative target quaternion of the robot B relative to the robot A, and note that $\bm{q}$ refers joint configuration of the robot, and $q$ represents quaternions.

To represent the motion time, a commonly used surrogate metric—also employed as a baseline in evaluations—is the distance between the starting and target joint configurations:
\begin{equation}
     \chi_{\text{mt}}(\bm{q}) =\| \bm{q} - \bm{q}_0 \|_2^2,
\end{equation}
where $\bm{q}$ is the joint configuration being optimized.
In this formulation, a bang-bang control strategy can be assumed at the velocity level, and the distance metric is normalized by introducing a weighting factor $w_i$ to account for the different velocity limits of each joint:
\begin{equation}
     \chi_{\text{mt}}(\bm{q}) = \sum_{i=1}^{n_T} w_i \left( q_i - q_{i,0} \right)^2.
     \label{eq:distance}
\end{equation}
where $n_T$ represents total number of \ac{DoF}s for the system.



\begin{figure}[t]
    \centering
    \includegraphics[width=0.45\textwidth]{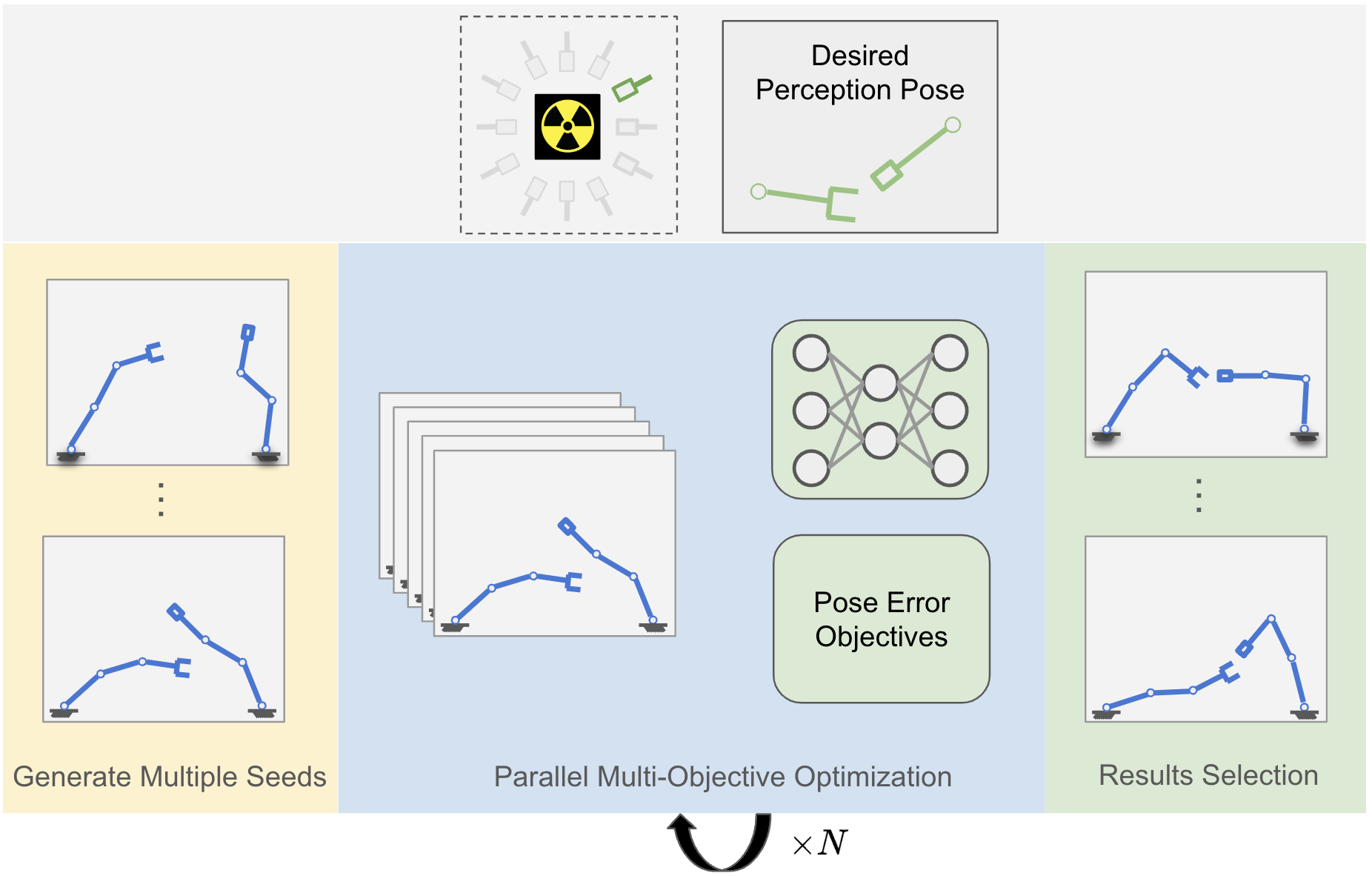}
    \caption{\textbf{Pipeline of ETA-IK}: given a desired perception pose, ETA-IK first generate a batch of initial configurations (yellow). A multilayer-perceptron time apporiximator are then integrated in a parallel optimization framework (blue). After $N$ iterations, the results is a batch of joint configurations. The best solution is selected according to different criteria (green).}
    \label{fig:pipeline2}
\end{figure}

\subsection{Execution Time Approximator}

\begin{figure}
    \centering
    \includegraphics[width=0.8\linewidth]{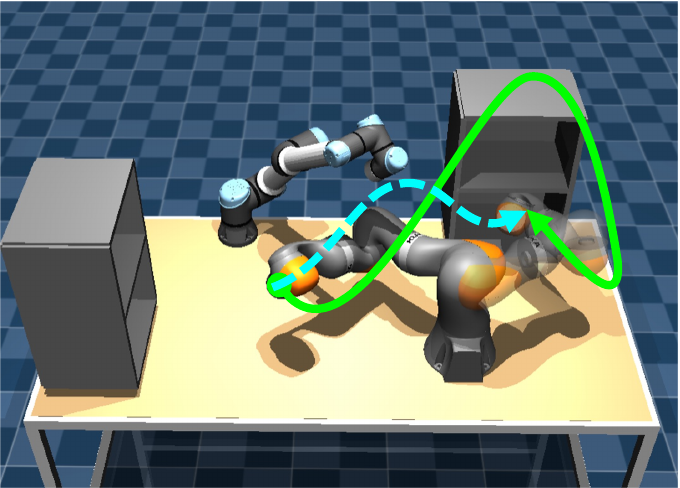}
    \caption{Execution time approximator is able to implicitly encode the self-collision or static environment information when the dataset considered collision avoidance. Considering the collision, the green trajectory is much longer than simply connecting the start and end configurations, shown in cyan.}
    \label{fig:why_time_estimator}
\end{figure}

Using surrogate loss such as distance between joint configurations as optimization objectives assumes that resultant \ac{PTP} motion between the start and target configuration is collision-free and executable. For a cluttered environment or a dual-arm setting, this assumption is usually not legitimate and therefore a post-processing step such as path planning is needed. This additional step introduce an gab between the distance-based surrogate loss and the actual execution time. Fig. \ref{fig:why_time_estimator} shows the difference of using cyan dashed surrogate distance and the green actual feasible trajectory. The actual trajectory is much longer and reflects a longer execution time. Furthermore, the robot's dynamic limitations, such as acceleration and jerk, also make surrogate distance metric impractical. Therefore, directly coping execution time in the optimization can provide a more timely efficient solution. 

For dual arms mounted on a fixed base, we can deploy offline optimal path planning or trajectory optimization algorithms to compute the collision-free motion between two arbitrary joint configurations and at the end acquire the actual execution time.
However, this paradigm can not be included into the multi-objective optimization due to its computation complexity. Instead, we collect data from these offline methods and use a multilayer perceptron (MLP) $f_\theta$ to approximate the execution time during optimization. Previous works \cite{huang2024planning} show that MLPs are capable of capturing the latent relation between joint configurations and time.

\textbf{Dataset}
The time approximator estimates the execution time of a collision-free trajectory between two configurations. It is worth mentioning that different trajectory generators can produce massively different results for the same start-goal query. For these reasons, two different generators, TOPPRA \cite{pham_new_2017} and Curobo TrajOpt \cite{sundaralingam_curobo_2023}, are used to collect datasets for training the time estimator. The key distinction between these two trajectory generators is that the Curobo TrajOpt takes collision into account and always returns collision-free trajectories, whereas the TOPPRA only considers the joint limits constraints. The dataset includes 250,000 collision-free start-target pairs and the respective execution time of the trajectories generated by TOPPRA and Curobo TrajOpt. 
In the dataset, approximately 50\% of the trajectories generated by TOPPRA are not collision-free, noting that all collisions in this context refer to self-collisions or collisions with static obstacles. In these cases, the execution time provided by TOPPRA serves a similar role as the surrogate distance, but it is more accurate since the dynamic limitations are considered.

\textbf{Training} 
The MLP takes the start-goal pair as input and outputs the estimated time. 
In practice, we use positional encoding $\bm{q}_{PE}$ = [$\bm{q}$, $\cos\bm{q}$, $\sin\bm{q}$] as input for the approximator. Given the start and target configuration $\bm{q}_0$, $\bm{q}_T$, the approximator outputs 
\begin{equation}
\chi_{\text{mt}}(\bm{q}, \bm{q}_0) = f_\theta(\bm{q}_{PE,0}, \bm{q}_{PE,T}).
\end{equation}
Literature \cite{koptev2022neural}\cite{mildenhall2021Nerf} suggests that this can improve the expressiveness of the MLP.
Both input and output is normalized to the range [-1, 1] based on the statistics of the dataset. It is worth mentioning that it is important that the learned MLP can correctly recolonize which start-goal pair has less execution time and provide the gradient of it. The absolute value of the estimation is not crucial in the use case. Throughout the training, mean square error (MSE) loss between the predicted $\chi_{\text{mt}}(\bm{q}, \bm{q}_0)$ and the ground truth in the dataset is used. The hyperparamters such as learning rate, numbers of the latent embedding are determined by a sweep process using Bayesian optimization.

\textbf{Inference} 
In the inference phase, namely the optimization phase, the trained MLP takes new configurations of every iteration as input and estimate the time. Since the optimization is a Quasi-Newton based method. The gradients of the MLP regarding the inputs are used for the optimization. MLPs can not only achieve fast inference time during optimization, but also support batch operations and provide gradients for the optimization.

\subsection{Optimization}
Inverse kinematics optimization aimed at achieving global optimality is computationally expensive. 
A recent study \cite{votroubek_globally_2024} reports an average computation time of 77.6 seconds to optimize a system with 10 DoF using the similar objective functions in \ref{subsec: multi-obj-ik}. 
Due to the curse of dimensionality, applying this method for online \ac{IK} is impractical.

The other non-convex optimization algorithms such as \cite{rakita_relaxedik_2018}, trade global optimality for potentially faster computation. To improve the chances of finding a better solution, a parallel optimization approach is often employed \cite{sundaralingam_curobo_2023}. This method uses multiple initial guesses in parallel, and the best result is selected as the optimal solution. Since the mapping relationship between Cartesian coordinates and the joints of a redundant manipulator is not unique, the number and distribution of initial guesses can significantly influence the final optimization result. 
The entire objective function is
\begin{equation}
    f(\bm{q}) = w_p \cdot \chi_p(\bm{q}) + w_o \cdot\chi_o(\bm{q}) + w_{mt} \cdot \chi_{\text{mt}}(\bm{q}, \bm{q}_0)+ w_b \cdot \chi_{b}(\bm{q})
\end{equation}
where $w_i$ represent the weight for the corresponding objective terms, and $\chi_{b}(\bm{q})$ is the joint limit cost proposed in \cite{sundaralingam_curobo_2023}.
Initially, Halton sampling is employed to generate samples from the configuration space that are free of collisions, serving as initial guesses for the optimization process. The step directions for the optimization are then computed based on the gradient of the defined objective function.


After several iterations, all optimization results are verified for feasibility and convergence to ensure that the calculated \ac{IK} solutions reside in a collision-free joint space and meet the predefined position and rotation error thresholds. A solution is considered successful if it satisfies both conditions. If a solution does not meet these criteria, an offset is added to the evaluation metric when selecting the best solution.

Typically, the \ac{IK} problem defines the optimal solution as the one that minimizes the pose error. Since the proposed method also consider the motion time, the metric for selecting the optimal solution can be defined in several ways: it can prioritize the pose error, the motion time, or a combination of both. This flexibility allows the optimization to balance accuracy in achieving the desired pose with the efficiency of motion, depending on the specific requirements of the task.

\section{Experimental Evaluations}

\begin{table*}[tbp]
    \centering
    \renewcommand{\arraystretch}{1.5}
    \setlength{\aboverulesep}{0pt}
    \setlength{\belowrulesep}{0pt}
    \newcommand{\est}[0]{{\color{blue} Est. Time }}
    \newcommand{\srr}[0]{{\color{purple} Surrogate Dist. }}

    \begin{tabular}{cl|c|c|c|c|c|c|c}
         \toprule
          & & Pick-IK-1 & Pick-IK-256 & \multicolumn{2}{c|}{ETA-IK-20} & \multicolumn{2}{c|}{ETA-IK-2k } & ETA-IK-200k\\ 
          \cline{5-9}
          & &   &  & \srr & \est & \srr & \est & \est \\
         \hline
         Execution Time & [s]  & 2.347  & - & 2.073 & \textit{2.055} & \textbf{1.963} & \textbf{\textit{1.926}} & \textit{1.892}\\
         Joint Configuration Dist.  & [$rad^2$] & 5.50 &  - & 4.71 & 4.73 & \textbf{4.33} & \textbf{4.33} & 4.28\\ 
         Computation Time & [ms] & 15.78 & 833.46 & \underline{59.40} & \underline{95.77} & \underline{62.48} & \underline{99.70} & 980.23 \\ 
         IK Success Rate & [\%] & 99.5 & 0 & 100 & 100 & 100 & 100 & 100\\
         \bottomrule
    \end{tabular}
    \caption{\label{tab:exp_3} Evaluation results based on random sampling from 10,000 IK instances. 
    Two variants of optimization objectives for ETA-IK are evaluated. While both of the variants use relative pose, \srr refers to the variant using joint configuration distance loss in the objective function and \est indicates the variant directly coping with the trained time approximator.  
    \textbf{Bold values} highlight that, even with similar joint distances, the execution time approximator identifies more optimal solutions.
    \textit{Italic values} show that increasing the number of optimization seeds further improves execution time.
    \underline{Underlined values} indicate that multi-threaded GPU inference ensures scalability without increasing computation time.
    }
\end{table*}

\subsection{General Evaluation in Simulation}
\label{sec: General Evaluation}
Initially, we configured a simple simulation environment featuring a dual Franka setup without environmental collisions to evaluate the general performance of the proposed method. 
This setup involves generating a random, self-collision-free initial joint configuration $\bm{q}_0$ and a random relative Cartesian pose $\bm{x}_{R,target}$.

For comparative analysis, we use Pick-IK (Reimplementation of Bio-IK \cite{8449979} by MoveIt2) as the baseline. The optimization includes joint configuration distance as an additional metric. We also evaluate multi-threaded computation on the CPU, which is provided as a built-in functionality of Pick-IK.
The evaluation criteria include:
\begin{itemize}
    \item The trajectory execution time after post-processing with TOPPRA
    \item The difference between the initial joint configuration and the computed solution
    \item The computation time required to solve the \ac{IK} problems
    \item The success rate of solving the \ac{IK} problems
\end{itemize}

For Pick-IK, the default setup with a single seed and CPU multi-threading with 256 seeds is selected as the baseline. The global mode is used for the \ac{PTP} \ac{IK} problem, and the \ac{IK} solver timeout is set to $0.5$ seconds. The weight for the joint displacement is set to $0.002$, as higher values result in an extremely high failure rate.

For ETA-IK, the default parameter values of Curobo \cite{sundaralingam_curobo_2023} are used. The joint configuration distance weights are set to $5$ and $20$ for the MPPI and Newton methods. The learned approximator applies displacement weights of $5$ and $50$, respectively.

The experiments were conducted on a desktop PC with an Intel i9 processor and an NVIDIA RTX 4090 GPU. The execution time approximator was also trained locally on this machine, requiring approximately 1 hour. Statistical results were evaluated over 10,000 independent inverse kinematics queries. Depending on the trajectory generation method used, the dataset generation time for 250,000 trajectories ranged from several hours to approximately one day.

\subsubsection{Execution Time Evaluation}
In general, \ac{IK} solutions obtained from the parallel multi-objective optimization framework result in shorter execution times. Even when the joint differences are identical, the execution time approximator can identify a more efficient solution for execution (see bold results in Table \ref{tab:exp_3}). Additionally, increasing the number of optimization seeds further improves the solution quality (see italicized results in Table \ref{tab:exp_3}).
\subsubsection{Computation Time Evaluation}
Although the proposed method generally exhibits longer computation times compared to Pick-IK, and the execution time approximator incurs additional inference overhead, it is important to note that the computation time remains unchanged when using 20 or 2000 seeds due to the multi-threading capabilities of the GPU (see underlined results in Table \ref{tab:exp_3}).
In contrast, Pick-IK with 256 threads experiences a significant computational overhead compared to its single-threaded counterpart. Moreover, the success rate of Pick-IK with 256 threads is zero, as all computations exceed the predefined time limit.

While the initial evaluation benchmarks the proposed method against baseline methods and standard metrics, the subsequent ablation study investigates how different design choices within the proposed framework affect performance, particularly in terms of shorter execution times and implicit collision handling.

\subsection{Ablation Study 
}
The real experimental setup for the dual-arm system, shown in Fig. \ref{fig:real_robot_experiment_setup}, consists of a UR5 and a KUKA LBR iiwa 14 R820 robot, with a combined total of 13 \ac{DoF}. The KUKA iiwa is designated for object grasping due to its higher load capacity, while a high-resolution 3D LiDAR Robin W from Seyond is mounted on the UR5's \ac{EE} for scanning purposes.

\begin{figure*}[t]
    \centering
    \includegraphics[width=0.8\textwidth]{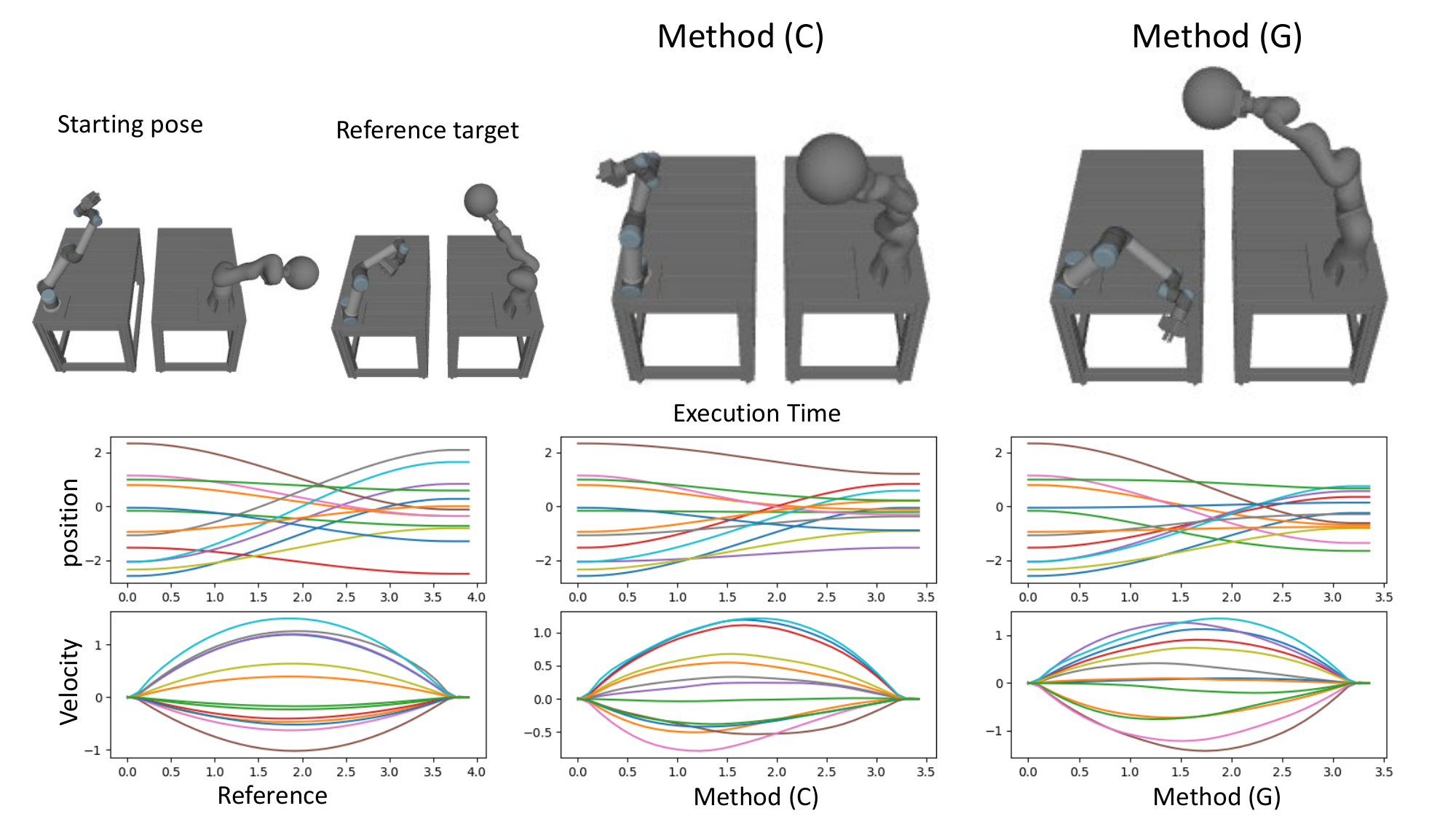}
    \caption{\textbf{Execution time comparison between reference, method (C) and (G) for collision-free trajectory generation}: The starting pose and the reference target 
 shown are randomly generated. The starting joint configuration, reference target configuration, and the IK solution generated by method (C) and (G) in Tab. \ref{tab:exp_2} are illustrated with the corresponding position and velocity profile over time generated by TrajOpt.}
    \label{fig:simulation}
\end{figure*}

In the simulation, additional static obstacles, such as the table, LiDAR, and viewpoint sphere, are present compared to the environment described in Section \ref{sec: General Evaluation} (as shown in Fig. \ref{fig:simulation}). To implement absolute \ac{TCP} pose \ac{IK}, we sample a random, collision-free initial joint configuration $\bm{q}_0$ and a random reference target joint configuration $\bm{q}_{target}$. The relative Cartesian pose $\bm{x}_{R,target}$ and the absolute \ac{TCP} poses $\bm{x}_{A,target}, \bm{x}_{B,target}$ of the two robots are then computed individually.

The robot joint limits used when generating the dataset for the network align with those in TOPPRA, including velocity $\left\{3.15, 3.15, 3.15, 3.2, 3.2, 3.2\right\}$ and $\left\{10.0, 10.0, 10.0, 10.0, 10.0, 10.0, 10.0\right\}$ $\frac{rad}{s}$, acceleration $\left\{5.0, 5.0, 3.0, 2.0, 2.0, 2.0\right\}, \left\{5.0, 5.0, 3.0, 2.0, 2.0, 2.0, 2.0 \right\}$ $\frac{rad}{s^{2}}$ constraints for UR5 and IIWA respectively. The maximum jerk for all the joints is $500\frac{rad}{s^{3}}$.

For the ablation study, different design choices are varied for evaluation. Methods (A) and (C) optimize Eq. (\ref{eq:distance}), using absolute and relative Cartesian poses as the surrogate execution time term in the objective function. Method (B) employs the trained execution time approximator but considers absolute Cartesian poses. Methods (D), (E), and (F) use the trained execution time approximator with different solution selection criteria, namely best time, best cost, and best pose error. All optimization parameters are summarized in Tab. \ref{tab:exp_1}. To explore the optimal solution, the number of initial guesses and optimization iterations are set at a relatively high value, namely $4096$ initial guesses and $120$ iterations. This ensures a comprehensive evaluation of the solution space while maintaining computation time within a practically acceptable range. The weights for position and orientation error are $2000, 2500$ and the execution time weights are set to be $250$ for joint distance and $500$ for execution time approximator.

\begin{table}[tbp]
    \centering
    \renewcommand{\arraystretch}{1.5}
    \setlength{\aboverulesep}{0pt}
    \setlength{\belowrulesep}{0pt}
    \begin{tabular}{@{}p{1.65cm}@{} p{2.1cm}@{} p{1.6cm}@{}|c|c|c}
        \toprule
        & & & Exec. T.  & Error  & SR  \\
        & & &  [s] & [mm] & [\%] \\
        \hline
        \multicolumn{3}{c|}{Reference $\bm{q}_t$} & 2.71  & 0     & -   \\
        \hline
        (A) Abs. TCP & + \color{blue} Surrogate Dist.  & & 2.618 & 0.4 & 100 \\
        (B) Abs. TCP & + \color{blue} Est. Time & + \color{violet} Best Time & 2.363 & 0.2 & 100 \\
        (C) Relative & + \color{blue} Surrogate Dist. & & 2.477 & 2.2 & 98  \\
        (D) Relative & + \color{blue} Est. Time & + \color{violet} Best Time & \textbf{2.053} & 1.6 & 100 \\
        (E) Relative & + \color{blue} Est. Time & + \color{violet} Best Cost & 2.450 & 0.3 & 100 \\
        (F) Relative & + \color{blue} Est. Time & + \color{violet} Best Pose& 2.807 & 0.3 & 100 \\
        \bottomrule
    \end{tabular}
    \caption{\label{tab:exp_1} Ablation study: execution time with random sampling from 100 IK instances using different input, objective functions and selection criteria. Execution time, position error and success rate (SR) are used as metrics. Specifically, absolute TCP and relative pose are two options for input desired pose. The objective options are marked in {\color{blue} blue} and the selection criteria are marked in {\color{violet} purple}.}
    \vspace{10pt}
\end{table}

\begin{table}[tbp]
    \centering
    \renewcommand{\arraystretch}{1.5}
    \setlength{\aboverulesep}{0pt}
    \setlength{\belowrulesep}{0pt}    
    \begin{tabular}{@{}p{1.0cm}@{} p{2.5cm}@{}|c|c|c|c}
         \toprule
          & & Exec. T. & Exec. T. & Error & SR\\
          & &  (TOPPRA) & (TrajOpt) & & \\
          & & [s] & [s] & [mm] &[\%]\\
         \hline
         \multicolumn{2}{c|}{Reference $\bm{q}_t$} & 2.483 & 3.522 & 0 & -\\
         \hline
         (C) Rel. & + Sur. Dist. & 2.4529 & 3.513 & 1.6 & 100 \\
         (G) Rel. & +  Est. T. (Col. free) & \textbf{2.394} & \textbf{3.40} & 2 & 100\\
         \bottomrule
    \end{tabular}
    \caption{\label{tab:exp_2} Resulting execution time with task-related relative pose from 100 IK instances. (C) Relative pose with distance loss; (G) relative pose with trained collision-free model.}
    \vspace{10pt}
\end{table}



\subsubsection{Execution Time Evaluation}
The execution time results indicate that using absolute Cartesian coordinates as position targets is less effective than relative positions for execution time optimization. The distance loss (C in the Tab. \ref{tab:exp_1}) and the proposed execution time approximator with cost based IK solution selection (E) demonstrate comparable performance in optimizing the execution time. It is worth noting that due to the greater optimization space, the positional error of (E) is significantly smaller. If the solution to success is sorted only by execution time cost, the solution obtained by proposed ETA-IK is almost $25\%$ shorter than the execution time of the randomly generated ref configuration, performs best among all comparison methods.

\subsubsection{Implicit Collision Consideration} 
For the approximator trained by the collision-free dataset, the \ac{NBV} candidate poses are utilized as target relative poses. Since these scanned positions are object-centered, they are more prone to collisions, thereby highlighting the influence of the collision-aware approximator on the \ac{IK} solution. As shown in the Tab. \ref{tab:exp_2}, the approximator trained with the collision-free dataset (G) demonstrates faster performance in both  \ac{TG} methods, with and without collision checking. Method (C), which employs relative position and distance loss as objectives, results in minimal joint changes (A typical example with random sampled starting and target is shown in Fig. \ref{fig:simulation}). However, due to the need for collision avoidance, the final collision-free trajectory execution time is longer than method (G). The symmetry of the velocity profile indicates the presence of obstacle avoidance behavior. While the reference trajectory is planned without encountering obstacles, resulting in a completely symmetrical velocity profile. The longer joint movement distances increased the execution time. In contrast, the IK solution obtained through the proposed approximator optimization exhibits relatively less obstacle avoidance behavior and achieves optimal execution time among the three methods considered.

\subsection{Real Robot Experiments}

The real robot experiments validated the motion generation pipeline, demonstrating that the generated trajectories perform effectively both in the simulation environment and on real robots. Videos demonstrating the system’s performance are available on the \url{https://yucheng-tang.github.io/eta-ik-website/}.


        

\section{Discussion and Conclusion}

In this work, we presented ETA-IK, an Execution-Time-Aware Inverse Kinematics method for dual-arm systems, with a specific application in scanning unknown objects during the dismantling of a nuclear facility. Our primary focus was on optimizing motion execution time by leveraging the redundancy of dual-arm systems. Unlike traditional trajectory generation techniques or physics-based simulators, ETA-IK leverages a trained execution time approximator that can be efficiently integrated into the optimization framework. This is enabled by its ability to support parallel inference and to provide differentiable outputs. 

Experimental results demonstrated that incorporating motion execution time directly into the IK optimization framework yields significant improvements in efficiency compared to traditional methods that rely on surrogate metrics, such as joint configuration distance. By incorporating collision avoidance into the optimization process, our approach could account for self- and static collision without introducing substantial computational overhead. This was particularly evident in both simulated and real-world scenarios, where the system was able to achieve more efficient execution while maintaining the relative pose between the two robots.

In addition, we identified a limitation of the proposed method: the additional inference time required by the neural network-based execution time approximator compared to simple distance-based surrogate metrics. However, this computational overhead remains nearly constant as long as the number of parallel threads does not exceed the capacity of the GPU. Nevertheless, it should be acknowledged that the proposed approach relies on access to a sufficiently powerful GPU to support efficient parallel inference and optimization within practical time constraints. Future research may explore reducing the overhead of the approximator further and learning more effective initial pose distributions—potentially replacing Halton sampling—with methods such as the one proposed in \cite{huang_diffusionseeder_2024}.

\bibliographystyle{IEEEtran}
\bibliography{bibliography/icra_eta_ik, bibliography/case2023, bibliography/supplement}

\end{document}